\documentclass[runningheads]{llncs}
\usepackage[T1]{fontenc}
\usepackage{graphicx}
\newcommand{\hourglassemoji}{\includegraphics[height=1.9ex]{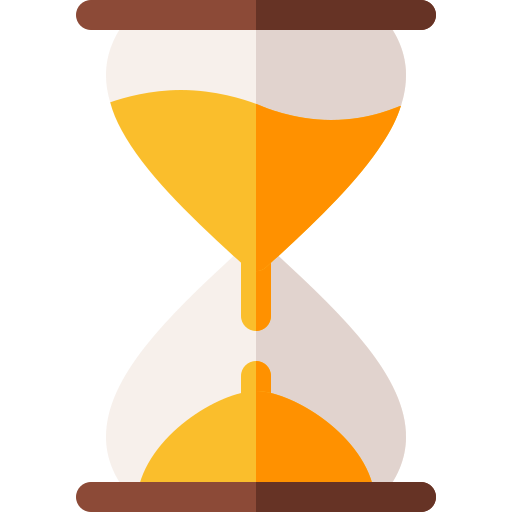}}

\usepackage{graphicx}
\usepackage{amssymb}
\usepackage{amsmath}
\usepackage{subcaption}

\begin{document}
\title{Temporal Fusion Strategies for Time-Aware Named Entity Recognition in Historical Texts}
\title{Ablating Time: Exploring Fusion Strategies for Temporal Representation in Historical NER}
\title{Telling Time: Temporal Fusion Strategies for Named Entity Recognition Across Centuries}
\title{\hourglassemoji\ A Study of Temporal Fusion Strategies for Named Entity Recognition in Historical Texts}
\titlerunning{Temporal Fusion Strategies for NER in Historical Texts}
% If the paper title is too long for the running head, you can set
% an abbreviated paper title here
%
\author{Emanuela Boros\orcidID{0000-0001-6299-9452}}
% \author{Anonymous}\inst{1}

\authorrunning{E. Boros}
% First names are abbreviated in the running head.
% If there are more than two authors, 'et al.' is used.

\institute{Digital Humanities Laboratory, EPFL, Lausanne, Switzerland}
% \\\email{emanuela.boros@epfl.ch}
%
% \institute{}
\maketitle              % typeset the header of the contribution
\begin{abstract}
Temporal variation poses a unique challenge for named entity recognition (NER) in historical texts, where entities drift in surface form and salience across time. While language models (LMs) have made progress in various NLP tasks, their ability to reason about temporality, especially in diachronic contexts, remains limited or at least, questionable. In this paper, we systematically study how temporal metadata can be structurally embedded into NER models using a range of lightweight fusion strategies. We experiment with both absolute and relative temporal representations, injected into Transformer-based architectures via early or late fusion mechanisms such as cross-attention, adapters, and concatenation. Our evaluations on French and German historical datasets reveal that late fusion strategies yield more robust and temporally generalisable performance, particularly in early and noisy periods.
\end{abstract}
\section{Introduction}

Language is inherently temporal: its vocabulary, structures, and referents evolve across time. Yet LMs, despite their generalization power, still struggle with temporal reasoning \cite{ding2024dolanguagemodels,jain2023dolanguagemodels,liu2024howcanlarge,liu2024stllmlarge,nako2025navigatingtomorrow,nylund2023timeisencoded,zheng2024understandingwhylarge,qiu2023arelargelanguage,wallat2024temporalblindspots,xiong2024largelanguagemodels}. Studies show that even advanced (generative) models like GPT-4 exhibit directionality biases \cite{papadopoulos2024arrowsoftime}, poor calibration over time \cite{beniwal2024rememberthisevent}, and difficulties retaining or reasoning over temporally anchored facts \cite{cole2022timeawarelanguage}. This limitation is particularly problematic in tasks such as named entity recognition (NER) over historical texts, where entities evolve, drift, or vanish entirely across time \cite{boros2020alleviating,ehrmann2020introducingtheclef,ehrmann2022introducingthehipe,ehrmann_extended_2022,pawłowski2024nlpfordigital,rijhwani2020temporallyinformed}. While temporality has been studied more commonly in video-based reasoning \cite{ko2023largelanguagemodels,liu2024stllmlarge}, in NLP tasks such as QA \cite{agarwal2018dianedtimeaware,chang2024acomprehensiveevaluation,gruber2024complextempqaalarge,jia2018tempquestionsabenchmark,ruiz2025onthetemporal,tan2023towardsbenchmarking,xiong2024largelanguagemodels}, or retrieval augmentation \cite{gade2024itsabout}, historical NER remains comparatively underexplored.

Recent research has introduced temporal representations like time vectors \cite{nylund2023timeisencoded}, timestamp-aware pretraining \cite{cole2022timeawarelanguage}, temporal graphs \cite{liang2022survey,lu2025knowledge,rosin2022time,song2023multilingual}, and dynamic knowledge editing \cite{yin2023historymatterstemporal} to help models encode temporal signals. Yet these remain largely disconnected from token-level tasks like NER. Interpretability studies such as probing \cite{gurnee2024languagemodelsrepresent,thukral2021probing} and temporal diagnostic tests like TEMPLAMA \cite{cole2022timeawarelanguage} confirm that temporal information is often only weakly represented in model weights.

In the domain of NER, earlier efforts to address time drift focused on sampling or data augmentation in high-churn environments like different platforms of social media \cite{chen2021mitigating,rijhwani2020temporallyinformed,ushio2022namedentityrecognition}. Meanwhile, historical NER introduces compounding challenges: diachronic drift, OCR degradation, and multilingual variation. Benchmarks such as HIPE \cite{ehrmann2020introducingtheclef,ehrmann2022introducingthehipe,ehrmann_extended_2022} have laid the groundwork, and newer work has begun exploring temporally aware grounding through context retrieval \cite{rosin2022time}, temoral knowledge graphs injection \cite{gonzalez-gallardo2023injectingtemporal}, or LLM-based inference \cite{hiltmann2025ner4allorcontext}. While this is a good start, none have systematically compared architectural fusion strategies or directly assessed in practice whether models internalise temporal information.

In this paper, we (1) systematically inject temporal information into a Transformer architecture using explicit year embeddings, (2) design and compare a suite of modular, interpretable fusion strategies that incorporate time at different points in the model (e.g., early vs. late), and (3) benchmark their impact across decades and languages, while probing whether the models genuinely internalise temporal signals. We hope that this study will contribute to a clearer understanding of how time can be structurally integrated into token-level models and inform future work in (practical) historical NLP and temporally-aware sequence modeling. %We hope that with this work, we provide a principled foundation for modeling temporality in historical NER, offering both practical improvements and theoretical insights into how time can be structurally integrated in sequence labeling tasks.

% Our paper addresses this gap by proposing a framework for temporal fusion in historical NER. Instead of relying on retrieval augmentation, prompting, or temporally stratified fine-tuning to simulate future prediction, we inject time directly into a Transformer-based architecture using various interpretable fusion strategies. %We consider both absolute year and relative (distance-based) encodings, benchmarking performance across two languages (French and German), decades, and fusion mechanisms (e.g. early, late). %Additionally, we design a probing setup that assesses whether temporal signals are truly internalized, by removing or perturbing year embeddings at inference time.

\section{Incorporating Temporality into NER}

\paragraph{Task Formulation.} We treat historical named entity recognition (NER) as a straightforward token classification task, just with a temporal twist. Each input consists of a sequence of tokens \( X = (x_1, x_2, ..., x_n) \), along with the document's publication year \( year \in \mathbb{N} \). The goal is to assign each token \( x_i \) a label \( l_i \), selecting from a standard entity taxonomy or marking it as non-entity. We use a Transformer-based architecture, where an encoder produces contextualized token representations \( H = \text{Encoder}(X) \in \mathbb{R}^{T \times d} \), with \( T \) the number of tokens and \( d \) the hidden size. Each label \( l_i \) is then predicted from \( h_i \in \mathbb{R}^d \), the contextualised representation of token \( x_i \).

\paragraph{Temporal Fusion Strategies.} To enable temporal adaptation of token classification models, we incorporate a temporal fusion module that integrates temporal context into token representations. This module fuses contextualised encoder outputs with a year-specific embedding using one of several strategies. We categorise them into \textit{two fusion types}: 
\begin{itemize}
    \item \texttt{early fusion}, where temporal information is injected before or during encoding; and 
    \item \texttt{late fusion}, where temporal information is applied to the encoder output.
\end{itemize} We explore these strategies in \textit{two modes of encoding temporal information}:
\begin{itemize}
    \item \texttt{absolute} mode, the embedding index corresponds directly to the publication year (e.g., 1889); and
    \item \texttt{time-distance} mode, we instead compute the number of years between the document's publication date and a fixed reference year, namely 2025, assigning lower indices to more recent documents.
\end{itemize}
More specifically, let $y = \text{Emb}({year}) \in \mathbb{R}^{d}$ denote the embedding of the document's publication year.

\paragraph{Baseline.}
This strategy skips temporal fusion entirely, i.e., $\tilde{H}_t = H_t$, and serves as a control condition.

\subsection*{Early Fusion}

\paragraph{Cross-Attention Fusion (early-cross-attention).}
Temporal information is injected \emph{before} encoding via cross-attention between the token embeddings and the year embedding:
\[
\tilde{H} = H + \text{MultiHeadAttention}(Q = H, K = y, V = y),
\]
where $H$ denotes the input token embeddings and $y$ is the year embedding, broadcast to match the input length. This mechanism allows each token to attend directly to the temporal context during encoding.

\subsection*{Late Fusion}

% \paragraph{Feature-wise Linear Modulation (FiLM).}
% A learned affine transformation modulates each token representation via year-conditioned scaling and shifting:
% \[
% \tilde{H}_t = \gamma(y) \odot H_t + \beta(y), \quad \gamma, \beta : \mathbb{R}^d \rightarrow \mathbb{R}^d.
% \]

\paragraph{Adapter Fusion (adapter).}
A lightweight MLP (adapter) processes the year embedding and adds the result to each token:
\[\tilde{H}_t = H_t + \text{MLP}(y), \quad \text{MLP}: \mathbb{R}^d \rightarrow \mathbb{R}^d.\] %

\paragraph{Concatenation Fusion (concat).}
Generic fusion technique in many tasks, The year embedding is concatenated to each token vector and projected back to the original dimensionality:
\[\tilde{H}_t = W \cdot [H_t; y], \quad W \in \mathbb{R}^{2d \times d}.\]

\paragraph{Relative Temporal Fusion (relative).}
A nonlinear encoder transforms the year embedding into a relative temporal representation, which is used in a feature-wise linear modulation (FiLM)-like modulation \cite{perez2018film}:
\[y' = \text{LayerNorm}(\text{SiLU}(W y)), 
\tilde{H}_t = \gamma(y') \odot H_t + \beta(y'), \text{where:}\]
\[\text{SiLU}(x) = x \cdot \sigma(x) \text{ is the sigmoid linear unit and } \sigma(x) \text{is the logistic sigmoid.}\]

\paragraph{Cross-Attention Fusion (late-cross-attention).}
Temporal information is fused with the encoder output using cross-attention, similar to the \textit{early fusion} one but \emph{after} encoding: 
\[\tilde{H} = H + \text{MultiHeadAttention}(Q = H, K = y, V = y).\]

\section{Experimental Setup}

\paragraph{Datasets.} Our experiments are based on the \texttt{hipe2020} dataset, as included in the HIPE-2022 shared task \cite{ehrmann_extended_2022}. We focus exclusively on the French and German subsets, which include publication year metadata required for temporal modeling (the English subset was excluded due to missing training data). We use the coarse-grained entity taxonomy (\textit{loc, org, pers, time, prod}) and retain all documents regardless of their temporal span. The French data comprises 10,923 annotated mentions across 1798--2018 with an average OCR noise rate of $\approx$33\%, while the German subset contains 6,584 mentions spanning 1798--1950 with $\approx$43\% OCR noise. While all splits cover wide temporal ranges, our goal is not to simulate chronological generalisation but to analyse the structural inclusion of time in the models.

\paragraph{Evaluation \& Hyperparameters.} We evaluate all models using micro-averaged F1 scores, computed at the entity level. All models are fine-tuned using the standard Transformer architecture, with the multilingual historical variant as base model\footnote{\url{https://huggingface.co/dbmdz/bert-base-historic-multilingual-cased}} \cite{schweter2022hmberthistoricalmultilingual}, with a maximum sequence length of 512 tokens. Models are trained using a batch size of 16, for 5 epochs, with a fixed seed (2025) for reproducibility.

%All experiments are run on GPU (CUDA-enabled), and temporal conditioning is implemented using a custom multitask architecture that optionally supports relative year encoding.

% To quantify temporal robustness, we further analyze variance in performance across time, comparing stability and generalization across early, middle, and late periods. All experiments are conducted using a shared transformer backbone with matched training conditions to ensure fair comparison. To comprehensively evaluate the temporal modeling capabilities of our entity recognition system, we designed a diverse set of quantitative and diagnostic analyses, targeting different aspects of performance, temporal awareness, and robustness. Our primary evaluation metrics are precision, recall, and F1 score, computed at the entity level. %We report these scores globally and disaggregated by:
% \begin{itemize}
%     \item \textbf{Entity type}: person (PER), location (LOC), and organization (ORG);
%     \item \textbf{Temporal bin}: based on decade, century, and finer year-based bins;
%     \item \textbf{Model strategy}: including baseline (no temporal adaptation), and multiple temporal fusion strategies (e.g., early and late cross-attention, relative positioning, adapters, concatenation, FiLM layers).
% \end{itemize}

\subsubsection{NER Performance Across Temporal Strategies.}

To evaluate the effectiveness of temporal conditioning, we plot F1 scores across publication years for each fusion strategy under two temporal modes: \texttt{absolute} and \texttt{time-distance} in Figure~\ref{fig:temporal-strategy-lines}. At first glance, we might not be able to see big improvements, but we do observe several slight temporal patterns across both languages:
\begin{figure}[ht]
    \centering
    \includegraphics[width=1\textwidth]{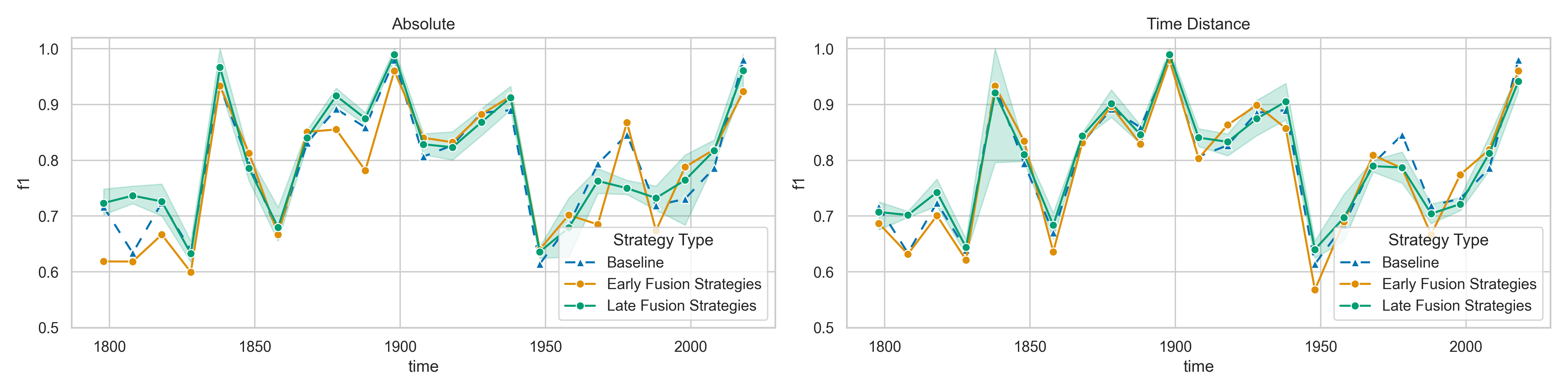}
    \includegraphics[width=1\textwidth]{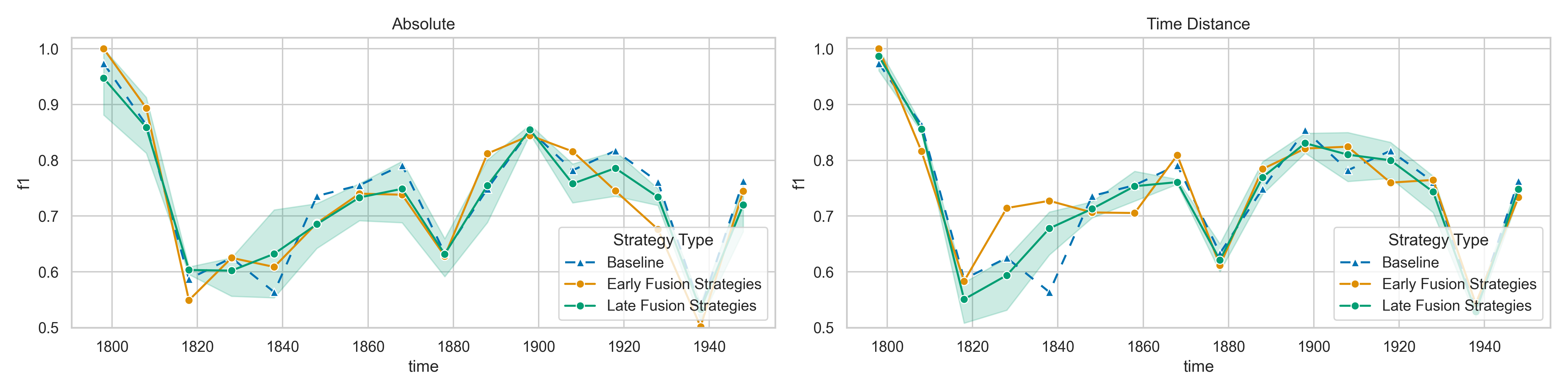}
    \caption{F1 scores over time for French (top) and German (bottom) subsets of HIPE-2020 under two temporal modes: \texttt{absolute} (left) and \texttt{time-distance} (right).}
    \label{fig:temporal-strategy-lines}
\end{figure}
\begin{itemize}
    \item \textbf{1800--1850}: Early periods exhibit high variability in F1 scores, likely due to OCR noise and sparse annotations. Late fusion strategies demonstrate notable gains in robustness, particularly under the \texttt{time-distance} mode, outperforming both baseline and early fusion.

    \item \textbf{1850--1900}: Performance stabilizes across models. While all strategies benefit from improved data quality, late fusion still maintains a slight edge, especially in French. Early fusion appears more sensitive to temporal encoding choices.

    \item \textbf{1900--1950}: F1 scores fluctuate again, particularly in German, with drops around 1940--1950. This may be attributed to document scarcity or inconsistencies in historical orthography. Late fusion again proves more resilient.

    \item \textbf{1950--2000}: Baseline models catch up, but late fusion strategies retain superiority, especially in the German subset. The performance gap narrows, suggesting a a a diminishing marginal benefit from temporal conditioning in modern decades.

    \item \textbf{2000--2018}: All models improve steadily due to better OCR and more standardised data. However, late fusion strategies still outperform slightly, reflecting their capacity to generalise across time even when temporal drift is lower.
\end{itemize}
Generally, we observe that all temporal fusion strategies, particularly late fusion ones, consistently improve NER performance across both languages with some benefits most pronounced in early or noisy periods, but before establishing the significance of these results, we analyse next other possible influencing factors. %, but they also remain competitive in more modern segments.

\subsubsection{Absolute Time versus Distance-based Encoding.}

\begin{figure}[h!]
    \centering
    \begin{subfigure}[t]{0.49\textwidth}
        \centering
        \includegraphics[width=\textwidth]{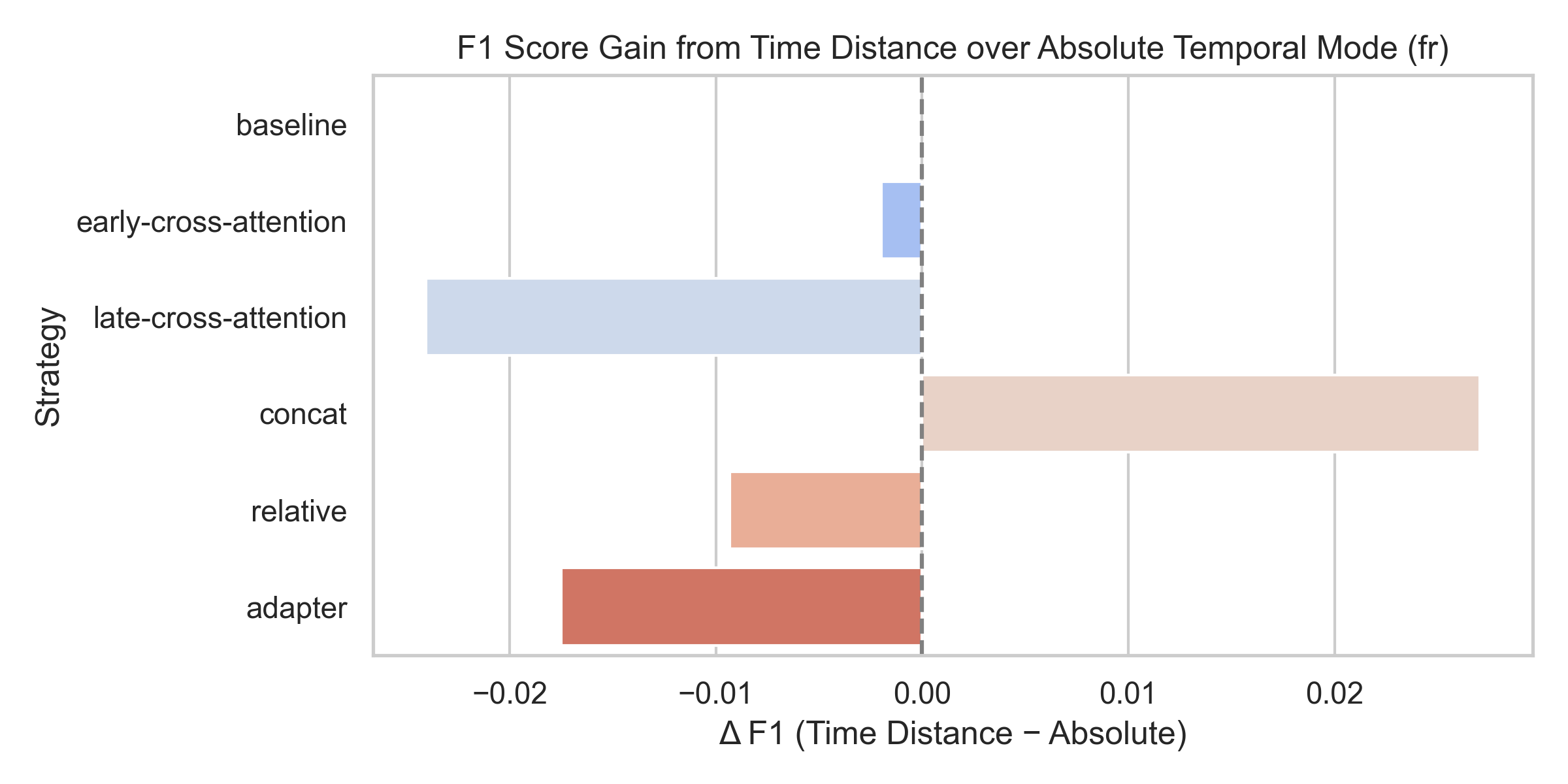}
        \caption{French}
        \label{fig:delta-f1-fr}
    \end{subfigure}
    \hfill
    \begin{subfigure}[t]{0.49\textwidth}
        \centering
        \includegraphics[width=\textwidth]{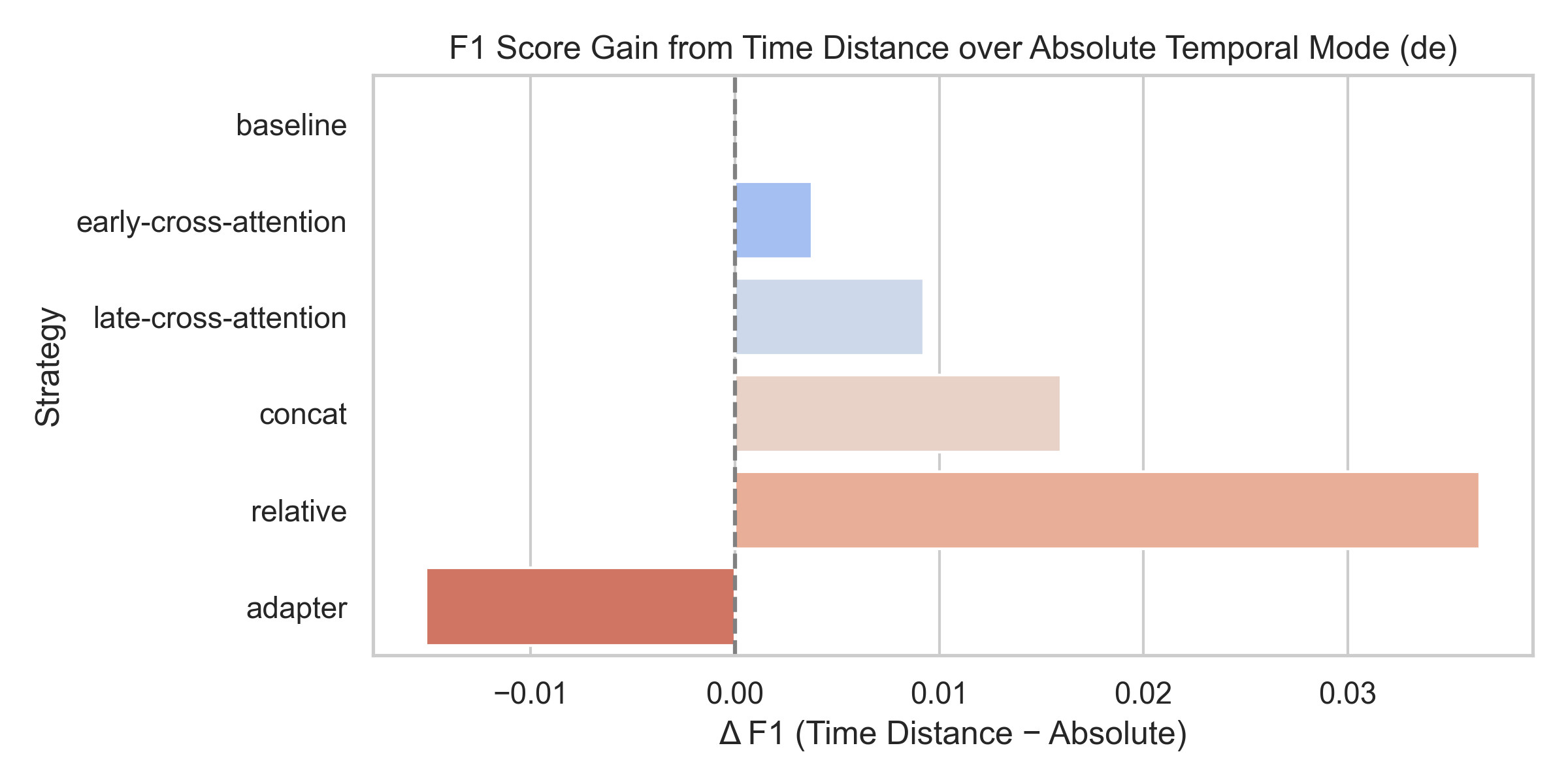}
        \caption{German}
        \label{fig:delta-f1-de}
    \end{subfigure}
    \caption{Average F1 score difference between \texttt{time-distance} and \texttt{absolute} temporal modes, computed for each fusion strategy. Positive values indicate improved performance.}
    \label{fig:delta-f1-temporal-mode}
\end{figure}

We compare the impact of \texttt{absolute} versus \texttt{time-distance} temporal encoding by computing the mean F1 score difference per strategy (Figure~\ref{fig:delta-f1-temporal-mode}). 
We see that, in German, strategies like \texttt{concat}, \texttt{relative}, and \texttt{adapter} benefit from \texttt{time-distance} encoding (up to +3 F1), suggesting improved temporal generalisation. In French, however, effects are less consistent: while \texttt{concat} gains slightly, others such as \texttt{adapter} and \texttt{late-cross-attention} perform better with \texttt{absolute} encoding. These results could imply that while the choice of temporal mode is secondary to the fusion strategy, it can still influence outcomes and should be tuned per language and setup.

\subsubsection{Entity Length Sensitivity.}

To explore whether temporal strategies differentially impact entity mentions of varying surface complexity, we categorize entities by their character length: those with 10 characters or fewer are considered \emph{short}, those between 11 and 20 as \emph{medium}, and those exceeding 20 characters as \emph{long}. For each group, we compute average F1 scores and analyze the performance gap between long and short entities, denoted as \(\Delta\)F1 = F1$_\text{long}$ - F1$_\text{short}$.
\begin{figure}[ht]
    \centering
    \includegraphics[width=\textwidth]{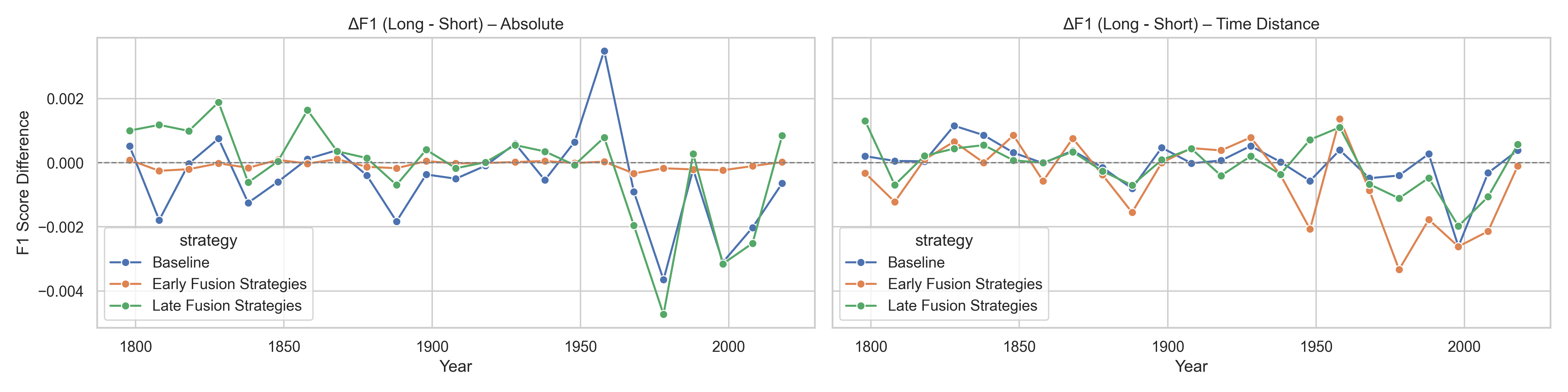}
    \vspace{0.5em}
    \includegraphics[width=\textwidth]{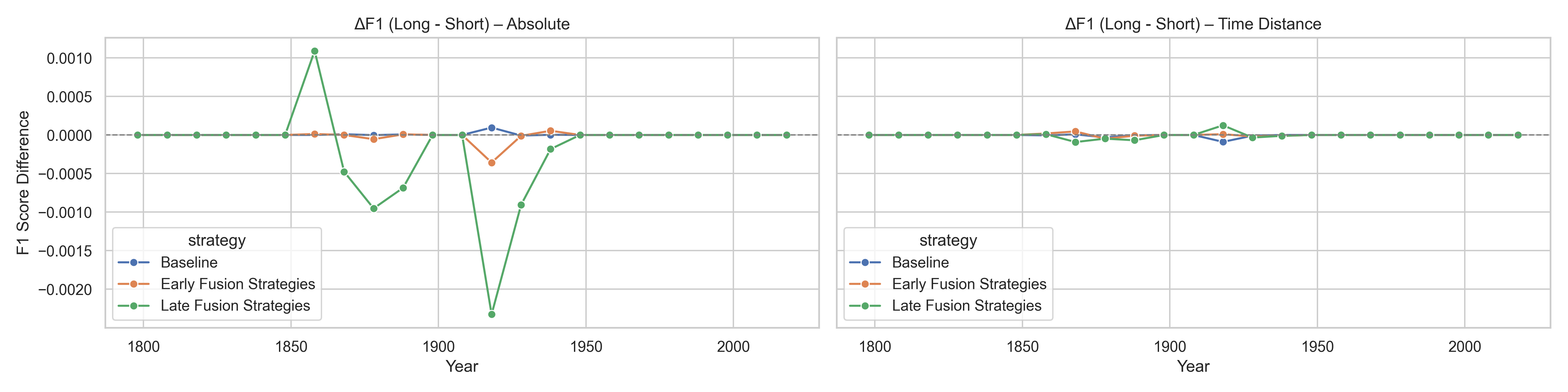}
    \caption{Difference in F1 score for French (top) and German (bottom) between long and short entity mentions (\(\Delta\)F1 = F1$_\text{long}$ - F1$_\text{short}$), across decades and fusion strategies for each temporal mode. }
    \label{fig:delta-length}
\end{figure}
Figure~\ref{fig:delta-length} presents the length sensitivity analysis across French (left) and German (right) subsets of the HIPE-2020 corpus. We observe that late fusion strategies tend to show a more stable or slightly positive gain for longer entities across both languages, particularly in earlier decades where surface forms tend to be longer or more structurally complex. The effect is more pronounced under the \texttt{time-distance} temporal mode, where relative temporal encoding appears to support generalisation over long spans. Baseline and early fusion strategies, by contrast, exhibit more unpredictability or minimal difference. These results suggest that injecting temporal signals at later stages of the model helps preserve surface-level distinctions critical for accurately identifying long entities.%, such as titles, compound names, or nested spans.

\subsubsection{Entity Type Gains from Temporal Fusion.}

\begin{figure}[ht]
\centering
\begin{minipage}{0.49\textwidth}
\centering
\includegraphics[width=\textwidth]{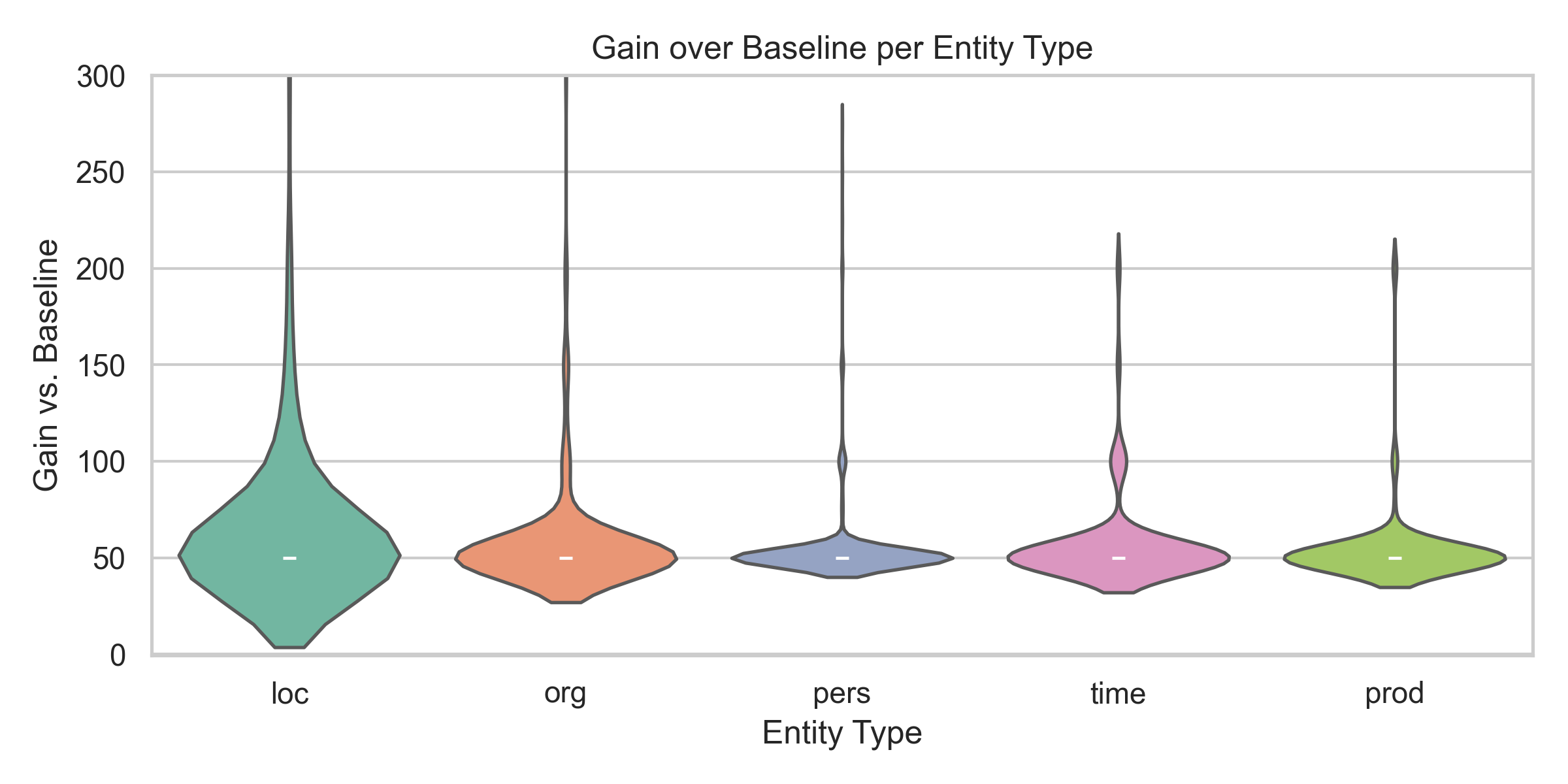}
\caption*{(a) French}
\end{minipage}\hfill
\begin{minipage}{0.49\textwidth}
\centering
\includegraphics[width=\textwidth]{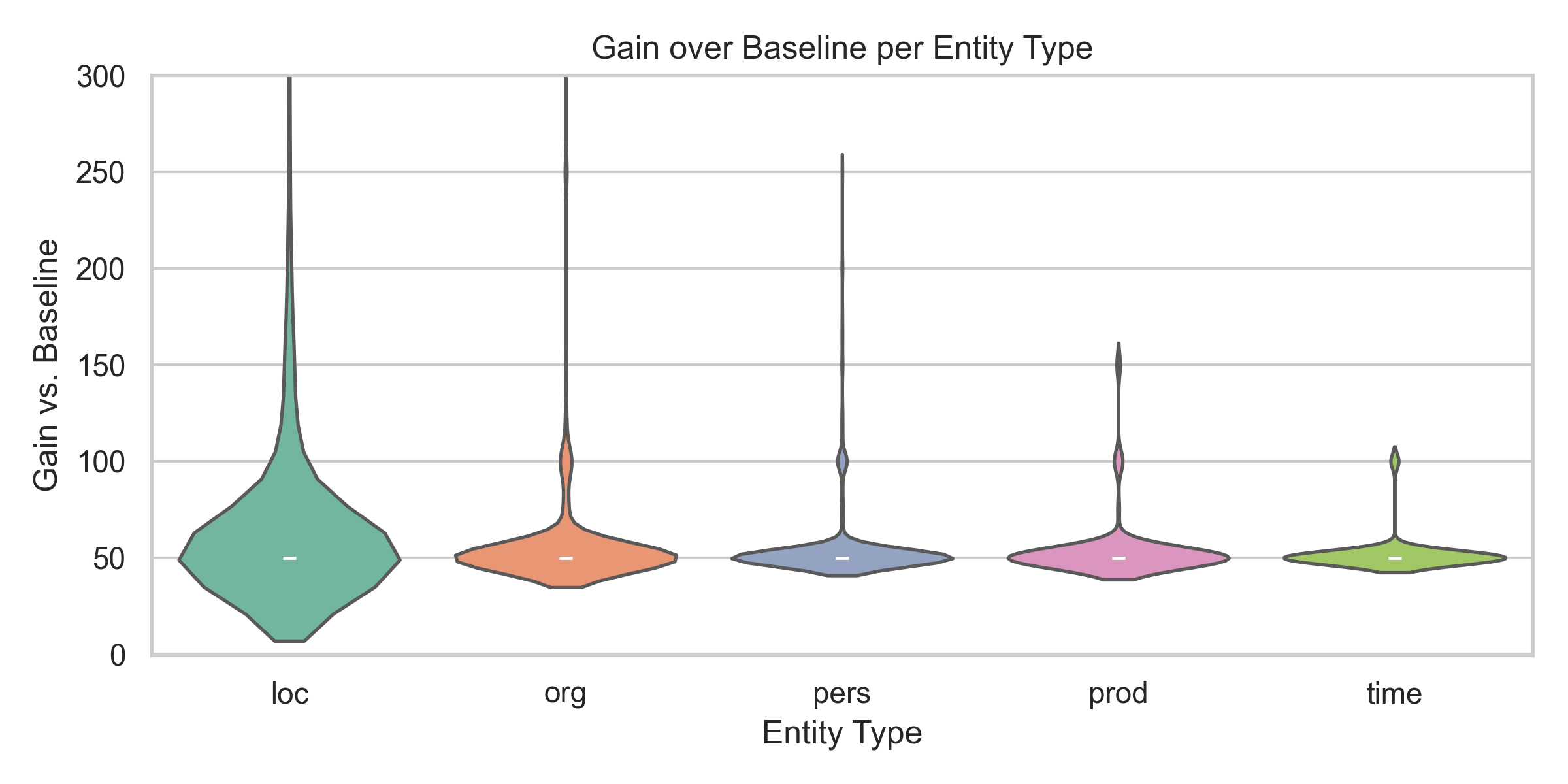}
\caption*{(b) German}
\end{minipage}
\caption{Distribution of gain over baseline for each entity type, measured as the difference in surface form frequency between temporal models and the baseline.}
\label{fig:entity-type-gain}
\end{figure}

To assess which entity types benefit most from temporal fusion, we compute the gain in prediction frequency for each surface–type pair, defined as the increase in count compared to the baseline. Figure~\ref{fig:entity-type-gain} shows that \texttt{loc} entities exhibit the highest variability and occasional large gains, suggesting that temporal conditioning improves their recall, likely due to historical drift and ambiguity. Other types such as \texttt{org}, \texttt{pers}, \texttt{prod}, and \texttt{time} show more modest and consistent distributions, indicating that improvements are generally limited in magnitude.

\subsubsection{Do the Time-based Models Really Learn Time?}
% \subsection{Probing Setup}
% - Extracting [CLS] embeddings, year classification
% - Accuracy per model
% - Correlation with NER performance
% \subsection{Interpretation}
% - Late fusion shows stronger temporal encoding
% - Architectural vs task-induced learning

To evaluate the extent to which our models encode temporal information internally, we adopt a {linear probing} strategy. %This involves training a lightweight classifier, specifically, a multinomial logistic regression model, to predict the publication year of a document based solely on the final-layer representation of the \texttt{[CLS]} token.
% \paragraph{Methodology.}
Let $h_{\text{CLS}} \in \mathbb{R}^d$ denote the final hidden representation of the \texttt{[CLS]} token. We train a linear classifier of the form: \[
\hat{y} = \arg\max_i \; W_i^\top h_{\text{CLS}} + b_i,\]
where $W \in \mathbb{R}^{d \times Y}$, $b \in \mathbb{R}^Y$, and $Y$ is the number of discrete publication years. To ensure that the probing task evaluates \emph{latent} temporal knowledge rather than reflecting direct access to input metadata, we modify the forward pass by injecting a randomly sampled publication year \( y \in \mathbb{N} \) during inference. This {disables architectural conditioning on the true document year} (whether \texttt{absolute} or \texttt{time-distance}). To account for randomness and obtain a more stable signal, we repeat the probing process five times and report the average accuracy across runs.
\begin{figure}[h]
    \centering
    \includegraphics[width=\linewidth]{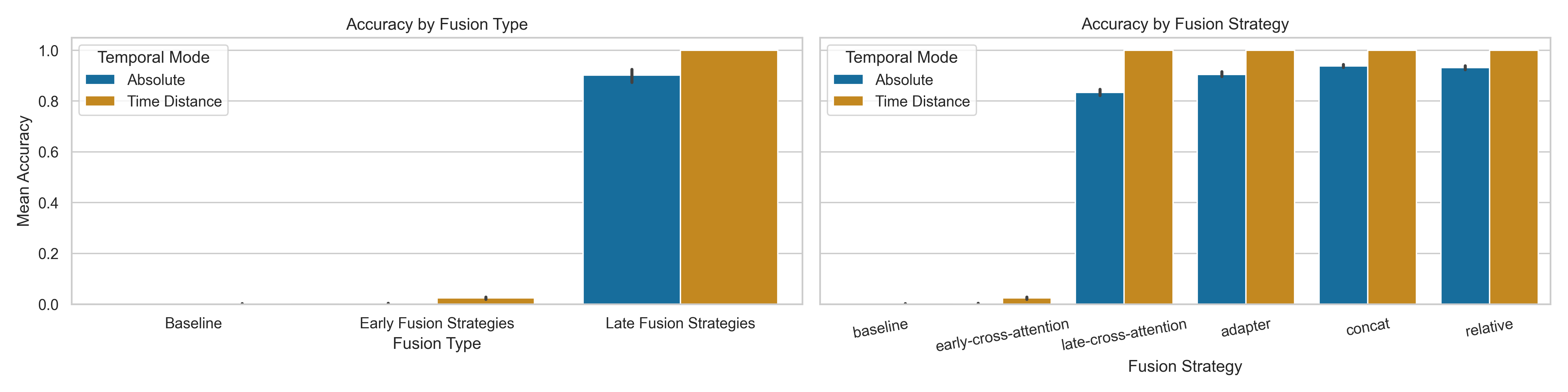}
    \caption{Probing accuracy across models. Left: grouped by fusion type. Right: grouped by fusion strategy.}
    \label{fig:probing_results}
\end{figure}
%\paragraph{Probing Results.}

Figure~\ref{fig:probing_results} reports the average prediction accuracy, grouped by fusion type and strategy, and we notice that the {late fusion strategies} consistently yield higher accuracy than early fusion or baseline models, confirming that injecting temporal information after contextualization better preserves temporal signals in the latent space. Among them, late cross-attention, adapter, and concatenation lead to the best results. In contrast, the baseline and {early-cross-attention} models show minimal temporal encoding. We could even say that probing shows that time-aware architectures, especially late-fusion models, encode temporality even when gold-year metadata is removed. This suggests that structural fusion mechanisms lead to genuine internalisation of temporal context, rather than relying on surface-level cues.

\subsubsection{Are the Improvements Really Significant?}
% To assess whether temporal fusion strategies lead to significant improvements over the baseline, we performed paired t-tests across yearly F1 scores for each strategy. Results reveal that, in most configurations, improvements are not statistically significant. 

% In the French subset (\texttt{fr}) with absolute temporal encoding, only the \texttt{late-cross-attention} strategy yields a statistically significant gain over the baseline ($p = 0.041$). All other strategies in this configuration, including \texttt{concat} ($p = 0.911$), \texttt{early- ross-attention} ($p = 0.172$), and \texttt{adapter} ($p = 0.770$), show non-significant results. Similarly, in the German subset (\texttt{de}), statistical significance is only achieved by \texttt{relative} encoding under the absolute temporal mode ($p < 0.001$), while all other strategies—including \texttt{late-cross-attention} ($p = 0.418$) and \texttt{concat} ($p = 0.650$)—remain non-significant. These findings suggest that although temporal strategies offer small gains, only a few yield statistically reliable improvements over the baseline.

To assess whether temporal fusion strategies offer statistically significant improvements over the baseline, we conducted paired t-tests across yearly F1 scores for each strategy and temporal mode. We noticed that most fusion strategies do not yield statistically significant gains at the $p < 0.05$ threshold with the exception of the \texttt{late-cross-attention} strategy under the \texttt{absolute} temporal mode demonstrates a significant difference compared to the baseline ($p = 0.041$). This could suggest that, while temporal fusion generally improves model performance, these improvements are often subtle and not uniformly consistent across years. %Nevertheless, the trend indicates that late fusion strategies—particularly those using absolute time encodings—have the highest potential for meaningful temporal adaptation.

\section{Insights \& Conclusions}

By structurally injecting time into Transformer-based architectures using modular fusion strategies, we demonstrate that temporal conditioning yields modest yet consistent gains for historical NER across languages, decades, and entity types. Late fusion strategies, particularly \texttt{late-cross-attention}, perform most robustly, especially in early, noisy periods, and help improve recognition of longer entities and temporally variable types like locations. Thus, based on our findings, we recommend: (1) adopting late fusion for integrating time; (2) testing both \texttt{absolute} and \texttt{time-distance} encodings, as their impact is context-dependent; and (3) using temporal fusion as a lightweight enhancement for diachronic or noisy corpora. While our approach is not novel and we acknowledge the growing utility of generative LLMs, we emphasize that real-world historical corpora often impose constraints: they may be large, private, or governed by restrictive policies. In such cases, structured methods that leverage metadata such as time or publication dates remain an important source of exploitable information for interpretable models.

\section*{Limitations}

While our study presents a systematic comparison of temporal fusion strategies for historical NER, there still remain several limitations. First, we only consider year-level granularity, which may be insufficient for domains requiring finer temporal resolution. Second, our experiments are confined to the HIPE-2020 dataset's French and German subsets, results may not generalise to other languages or genres of historical text. Third, the probing analysis focuses solely on linear decodability of year embeddings and may underestimate more subtle forms of temporal encoding. Finally, our models are evaluated under controlled conditions using a single backbone architecture; real-world applications with noisy or missing metadata may yield different results.

\bibliographystyle{splncs04}
\bibliography{mybibliography}
%
% \begin{thebibliography}{8}
% \bibitem{ref_article1}
% Author, F.: Article title. Journal \textbf{2}(5), 99--110 (2016)

% \bibitem{ref_lncs1}
% Author, F., Author, S.: Title of a proceedings paper. In: Editor,
% F., Editor, S. (eds.) CONFERENCE 2016, LNCS, vol. 9999, pp. 1--13.
% Springer, Heidelberg (2016). \doi{10.10007/1234567890}

% \bibitem{ref_book1}
% Author, F., Author, S., Author, T.: Book title. 2nd edn. Publisher,
% Location (1999)

% \bibitem{ref_proc1}
% Author, A.-B.: Contribution title. In: 9th International Proceedings
% on Proceedings, pp. 1--2. Publisher, Location (2010)

% \bibitem{ref_url1}
% LNCS Homepage, \url{http://www.springer.com/lncs}, last accessed 2023/10/25
% \end{thebibliography}
\end{document}